%% file: dissertation.tex
\documentclass[]{dukedissertation}
\usepackage{amsmath, amssymb, amsfonts, amsthm}
\usepackage{graphicx}
\usepackage[numbers]{natbib}
\usepackage{float}
\usepackage{color}
\usepackage{bm}
\usepackage{subfigure}
\usepackage{graphicx}
\usepackage{mathabx}
\usepackage{multirow}
\usepackage{setspace}
\usepackage{lineno}
\usepackage{paralist}

\graphicspath{{./Pictures/}}

\author{Robert Frank Martorano III}

\title{Semantic Term ``Blurring'' and 
 Stochastic ``Barcoding" for Improved Unsupervised Text Classification}

\supervisor{Xiaobai Sun}
\department{Computer Science} 

\date{2018} 

\copyrighttext{%
All rights reserved except the rights granted by the 
\\ 
   \href{http://creativecommons.org/licenses/by-nc/3.0/us/}
        {Creative Commons Attribution-Noncommercial Licence}
}

\member{Nikos P. Pitsianis}
\member{John H. Reif}

\usepackage[pdftex, pdfusetitle, plainpages=false, 
				letterpaper, bookmarks, bookmarksnumbered,
				colorlinks, linkcolor=black, citecolor=black,
	         filecolor=black, urlcolor=black]
				{hyperref}

\begin{document}
\doublespacing

\input{./macros.tex}

\maketitle

\include{{./Abstract/abstract}}

\tableofcontents 
\listoftables	
\listoffigures	


\include{{./Acknowledgements/acknowledgements}}

%
%
%
\include{{./Introduction_P1/introduction}}

\include{{./Overview_P2/overview}}

\include{{./Term_Blur_P3/term_blurring}}

\include{{./S_Barcoding_P4/s_barcoding}}

\include{{./Experimental_results_P5/experimental_results}}

\include{{./Discussion/discussion}}

\nocite{*}
\bibliographystyle{plainnat} 
\cleardoublepage
\normalbaselines 
\addcontentsline{toc}{chapter}{Bibliography} 
\bibliography{./Bibliography/References}

\include{{./Biography/biography}}

\end{document}

%% file: macros.tex

\newcommand{\dt}{$\mbox{Documents} \times \mbox{Terms}$ }
\newcommand{\dtNoS}{$\mbox{Documents} \times \mbox{Terms}$}
\newcommand{\dtSmall}{$\mbox{DT}$ }
\newcommand{\dtSmallNoS}{$\mbox{DT}$}

\newcommand{\dse}{$\mbox{Documents} \times \mbox{Semansic-Element}$ }
\newcommand{\dseNoS}{$\mbox{Documents} \times \mbox{Semansic-Element}$}

\newcommand{\dseSmall}{$\mbox{DSE}$ }
\newcommand{\dseSmallNoS}{$\mbox{DSE}$}

\newcommand{\barcode}{$\mbox{S-Barcode}$ }
\newcommand{\matlab}{$\mbox{MATLAB}$ }

%% file: Abstract/abstract.tex
\abstract
The abundance of text data being produced in the modern age makes it increasingly important to intuitively group, categorize, or classify text data by theme for efficient retrieval and search. Yet, the high dimensionality and imprecision of text data, or more generally language as a whole, prove to be challenging when attempting to perform unsupervised document clustering. In this thesis, we present two novel methods for improving unsupervised document clustering / classification by theme. The first is to improve document representations. We look to exploit ``term neighborhoods'' and ``blur'' semantic weight across neighboring terms. These neighborhoods are located in the semantic space afforded by ``word embeddings.'' The second method is for cluster revision, based on what we deem as ``stochastic barcoding'', or ``S-Barcode'' patterns. Text data is inherently high dimensional, yet clustering typically takes place in a low dimensional representation space. Our method utilizes lower dimension clustering results as initial cluster configurations, and iteratively revises the configuration in the high dimensional space. We show with experimental results how both of the two methods improve the quality of document clustering. While this thesis elaborates on the two new conceptual contributions, a joint thesis by \citet{david2018}, details the feature transformation and software architecture we developed for unsupervised document classification.


%% file: Acknowledgements/acknowledgements.tex
\acknowledgements
Dr. Xiaobai Sun for pushing me to explore areas I was unfamiliar, and having early confidence that I could reach levels I did not think possible. I started my Duke career in your ``Brain Mapping'' seminar, and you have been the most influential person in my studies here at Duke. My independent study work with you in mobile health data, ``MobiMed'', inspired me to work at Verily on nearly the exact same project after graduation.  I can't imagine how different my 4 years at Duke would look without your mentorship.

Other professors and academic mentors and advisors who have pushed me tremendously throughout my 4 years at Duke: Dr. Nikos Pitsianis, Alexandros Iliopoulos, and Tiancheng Liu.

All family and friends especially: Rob Martorano, Eileen Martorano, Tara Martorano, and Lisa Sapozhnikov.

%% file: Introduction_P1/introduction.tex
\chapter{Introduction}
Text classification is a vital, yet challenging undertaking. Technology has significantly increased the rate at which we gather text data, and has in great part led to an abundance of information to be consumed. Thus, it is more important now than ever to be able to categorize and group text data based on topic and content. This is a particularly prevalent issue in academia, where the rate of research and publication is on the rise, and keeping up with the trove of new information only becomes increasingly more challenging. Our particular focus is on two areas of text classification: (1) the representation of documents in a high dimensional term feature space, and (2) the manner by which the documents are organized into theme-based categories. These two areas are extremely important for any form of unsupervised learning with the objective of classification / clustering. In traditional data science, these are two explicit and basic steps in a pipeline we will discuss later. 

We chose to focus on unsupervised clustering for a variety of domain specific reasons. First, supervised classification requires pre-defined categories. In the domain of academic corpora, areas of research and ``subjects'' are always in flux. Much of the best academic work is interdisciplinary, and we believe pre-defined ``classes'' pigeonhole documents into sub-optimal groupings, based on the past (and potentially outdated) state of research. The primary reason for categorization of work and research is for efficient exploration and retrieval. We have become accustomed to keyword searching in our pursuit of finding knowledge, which while effective in pairing with categorization, we believe should not be the only option. In the work of \citet{kiwi2017}, they introduce the concept of a seed-paper search, a method in which a user can input a few papers of interest and explore the citation graph (bi-directional), and its respective content, to discover more relevant material. When we begin to explore academic work in a more visual manner, clustering of work can instead be executed based on theme, and connections can be naturally discovered, rather than forcing binning into pre-defined categories. Thus, we look to improve these theme clusters, that do not require human labeled training data and pre-defined areas of knowledge, but instead are based on natural associations between documents that are discovered through their respective usage of language. A final note: unlike other areas of text classification that often contain additional metadata which can be helpful in classification, the only information available in bulk for quick retrieval among academic papers is typically: the author, the title, and the abstract.


%

%% file: Overview_P2/overview.tex

\chapter{Overview}
\label{chap:overview} 

While variation may exist, a typical unsupervised pipeline for data
clustering consists of the following stages: 
\begin{inparaenum}[(1)] 
\item data acquisition or collection; 
\item data curating or filtering; 
\item feature dimension reduction; 
\item clustering in a dimension reduced space; 
\item post analysis and evaluation 
\end{inparaenum} 

I will first give an overview of a few influential and effective methods for
dimension reduction and clustering. I will then introduce ``word
embeddings", which we utilize for improved document feature
representation, prior to dimension reduction.

\section{Dimension Reduction}
\label{sec:dimension-reduction} 

Text data are naturally first expressed by the terms used in the text,
which results in an extremely high dimensional feature space. The
conventional way of representing a document is as a row vector of term
occurrences or occurrence frequencies. A document corpus is thus
represented by a large table or matrix \dtNoS, \dtSmallNoS. In \dtSmallNoS,
each row vector corresponds to a document in the corpus, each column
corresponds to a term used in at least one of the documents. 
The first step in a text classification pipeline is reducing the
dimension of the document features. There are two basic types of 
feature reduction. The first type is feature selection, where some terms 
are to be filtered out. The second type involves transforming the real-word features into synthetically encoded feature representation. 

The first operation for term selection is to remove ``stop words'',
which are extremely common terms that provide little value to information
differentiation. Stop words include ``the'', ``and'', ``in'', etc. We additionally filter through explicit
thresholding, removing the most and least frequently used terms
throughout the corpus. These thresholding
operations can be thought of as high and low pass filters.

Following selection, we then perform a feature transformation.  The
objective is maximize discernment between transformed document features
in an embedding space with a minimal number of dimensions. There are two
main categories for transformative dimension reduction techniques:
\begin{inparaenum}[(a)] 
\item Graph spectral embedding. This method casts the relationship
  between feature vectors as a graph first, and attempts to preserve
  geometric relationship among the feature vectors. A sparse algorithm
  for Singular Value Decomposition (SVD), made possible by the the work of \citet{Baglama} and \citet{bidiag}, is typically used
  for spectral embedding with a small number of singular vectors as the
  axial vectors. 
  
  \citet{Rege2006} used a bipartite to describe the relationship between
  documents and terms, use SVD to obtain the bipartitie
  spectral decomposition and then make c-clustering among documents and
  terms simultaneously. Long before, SVD was used for Latent Semantic
  Analysis (LSA) by \citet{lsa}.

\item Stochastic neighborhood embedding (SNE). This method casts
  local geometric relationship among feature vectors into conditional
  probability distributions, and then attempts to preserve the stochastic 
  neighborhood relationship. 
   A particular algorithm is t-SNE from \citet{VanDerMaaten2008}, which built off of the precursor work of SNE by \citet{Hinton2002}. 
\end{inparaenum}

\subsection{Bipartite Spectral Embedding}

In the paper, \citet{Rege2006} represents the relationship
between documents and terms as a bipartite graph,
$G(V_{doc}, V_{term},E, W_e)$, where the edge $e$ between a document and
a term exists if and only if the term is used by the document, the edge
weight $w_e$ is the count of the times the term is used.  This forms the
sparse \dt matrix, \dtSmall.

The adjacency matrix for the bipartite $G$  is 

\begin{equation} 
      A = \left(  \begin{array}{cc} 
                0      &  \dtSmall 
          \\ 
            \dtSmall^{\rm T}  & 0 
          \end{array}   \right) 
\end{equation} 
Dhillon shows how to relax the ``min cut" of the particular graph
into a generalized eigenvalue and eigenvector problem.  The cut can be
indicated by the second eigenvector, corresponding to $\lambda_2$, of
the graph laplacian, $L=L(G) = D-A$, where $D$ is the diagonal degree matrix. In the case of a bipartite:
\begin{equation} 
    D = \left(  \begin{array}{cc} 
              D_1      &  0
        \\ 
          0  & D_2
        \end{array}   \right) 
\end{equation} 
where $D_1(i,i) =  \sum_{j}{DT(i,j)}$ and $D_2(j,j) =  \sum_{i}{}{DT(i,j)}$.
In order to efficiently calculate the eigenvector corresponding to
$\lambda_2$, Dhillon shows how the singular value decomposition (SVD) of
the matrix $\dtSmall$ (term frequency matrix). This is shown below:

$$L \times z = \lambda \times D \times z$$
    $$
        \begin{bmatrix}
            D_1 & -\dtSmall \\
            -\dtSmall^T & D_2
        \end{bmatrix}      \begin{bmatrix}
            x \\
            y
        \end{bmatrix}
        =
        \lambda 
        \begin{bmatrix}
            D_1 & 0 \\
            0 & D_2
        \end{bmatrix}
        \begin{bmatrix}
            x \\
            y
        \end{bmatrix}
    $$
    $$ D_1^{-1/2} \times \text{\dtSmall} \times D_2^{-1/2} \times v=(1-\lambda) \times u $$

    $$D_2^{-1/2}  \times \text{\dtSmall}^T \times D_1^{-1/2} \times u=(1-\lambda) \times v $$

    Which is the same as the SVD of the normalized \dtSmall
    $$\text{\dtSmall} \tilde = D_1^{-1/2} \times \text{\dtSmall} \times D_2^{-1/2}$$
For two-cluster analysis, the document singular vector will suffice. For multipartition, or many-cluster analysis, more singular vectors are used. That is, the dimension of the spectral embedding and encoding space is greater.
In his paper, Dhillon uses the popular clustering algorithm K-means to find the clusters post dimension reduction. This method has the drawback of requiring knowledge of roughly how many clusters exist, or must used many times with a different number of expected clusters.
\subsection{Latent Semantic Analysis}
Latent Semantic Analysis (LSA) by \citet{lsa} was a breakthrough paper at the time and still is one of the most influential papers in Natural Language Processing. LSA focuses on the relationship between documents and terms, working with the \dt matrix. In comparison to Dhillon's method there are there are three important differences: (1) the paper makes no direct connection  graph analysis, (2) a different weighting scheme is used for the transformed features, and (3) no connection to clustering is made. In LSA the feature axes are scaled by their corresponding singular values. The basic idea of Latent Semantic Analysis is the low rank approximation of the \dtSmall matrix. It does not explore downstream clustering, and is limited to learning inter-corpus document-term relationships. We look to alleviate this problem.
\subsection{t-Distributed Stochastic Neighborhood Embedding}
Stochastic Neighborhood Embedding (SNE), was introduced by \citet{Hinton2002}, and was followed by t-SNE from \citet{VanDerMaaten2008}. Unlike many other dimension reduction methods like Principal Component Analysis (PCA), ISOMAP, or Locally Linear Embedding (LLE), which all focus on maintaining distance between dissimilar observations, t-SNE focuses on maintaining near neighbor relationships in the lower dimensional space. The method begins by converting the high dimensional distances into conditional probabilities. The technique then looks to minimize the  Kullback-Leibler divergence between the conditional probabilities in the high dimensional original feature space and the corresponding conditional probabilities in the desired lower dimensional feature space. It computationally reaches the best match by using gradient descent. t-SNE utilizes a student t-distribution in the low dimensional space, while SNE used a gaussian distribution. In comparison to SNE, t-SNE better preserves neighborhoods at multiple scales. 

t-SNE's improved near neighbor relationships in comparison to other methods come at the cost of not preserving geometric relationships. This causes the technique to incorrectly unroll the classic dimension reduction test data example, the swiss-roll, which methods like ISOMAP, can easily unroll.
For applications where neighborhood preserving is important, t-SNE does have the advantage of creating more appealing, or enlightening visualizations of the data, with greater separation between clusters in the lower dimensional space. This makes it easier for clustering techniques, such as mean shift to properly identify the clusters in the lower dimensional data. Again though, the geometry in the lower  dimensional embedding space is not necessarily a good approximation to that in the original space.

\section{Clustering}
Clustering is the process by which data points are separated into groups, where the intra-group data points are highly similar and data points in different groups are dissimilar. We used two clustering algorithms: mean shift and K-means; each requires a single parameter.
\subsection{Mean shift}
Mean shift is a commonly used non-parametric clustering algorithm, from \citet{Comaniciu2002}, with cluster formation determined by estimated density. The algorithm views the points in a feature space as an empirical density function (the particular density function can be specified), and looks to find the local maxima (or modes) of the distribution. In order to find these maximas, the algorithm treats each data point as a the center of a window, and calculates the mean of each window (window size based on bandwidth parameter). The window is then shifted to the mean of the window and this process is repeated until convergence. Points with similar convergence locations are merged.
\\Mean shift has the benefit of being non-parametric, meaning it does not assume a particular distribution of the data. It also does not require an input of the expected number of clusters. It, however, does require a non-trivial, and quite sensitive bandwidth parameter, which greatly impacts the  clustering results. It also does not scale well to high dimensions, as the geometry it relies on begins to fall apart. This is why it would be unwise to attempt to cluster the raw document representations in the original extremely high dimensional space.

\subsection{K-Means}
K-means is also a non-parametric clustering algorithm, from \citet{HartiganJ.A.andWong1979}, which looks to partition data points into K clusters. K-means looks to find K centroids, and decides which cluster a point ``belongs'' to by finding the centroid which is closest to the particular point. The value of K, is specified by the user. The distance metric can be specified by the user. Commonly, euclidean distance is used. K-means begins by randomly initializing the K centroids. For each point, it assigns it to the ``cluster'' corresponding to the closest centroid. It then calculates a mean value for each cluster and uses those means as the new centroid values. Once those means converge the process is over.\\
Even though K-means is non-parametric, needing to choose the K value requires some form of prior knowledge about the distribution of the data. While with mean shift the number of clusters is often sensitive to the bandwidth parameter, with K-means the number of clusters is actually explicitly assigned by the user. K-means is also sensitive to centroid initialization and to outliers. This is problematic, as if an outlier becomes its own cluster, it may go against your expected number of clusters, K, that was originally provided.

\section{Feature Transformation}
Here, feature transformation refers to a transformation executed prior to dimension reduction. The goal of a feature transformation is to improve the representation of the data. In our case, we utilize ``word embeddings'', which I will introduce below, to essentially perform a feature expansion that transforms our features into more associated, and stronger descriptors of our documents.
\subsection{Word Embeddings}
One of the major challenges with linguistics is that there was no intuitive geometric space in which language can be modeled, unlike a 2D color image, which falls on a manifold in low-dimensional parameter space, with 2 spatial dimensions, 3 color intensities and other optical and stochastic parameters. Language is complex, diverse, and evolving. This evolution occurs for a variety of reasons. For one, language is evolving to fit the needs of the user. Language is extremely low bandwidth and imprecise. We use the term ``semantic'' to describe the meaning of language, but the challenge with language, unlike an image again, is that the only way we can define language is using more language, unlike an image where we can define the construct using numerical methods. 

To combat these issues, researchers have attempted to create a “pseudo” feature space to describe language, at the granularity of words, in digital and vector representations. This is refereed to as ``word embeddings''. While researchers over the past 20 years have had varying success, there was a recent breakthrough in the area of ``word embeddings'', with the creation of word2Vec by  \citet{Mikolov2013}. Due to computational progress and the advancement of neural networks, the research team was able to create a semantic feature space to describe words, with two key properties: near neighbor relationships between words with similar semantics and rudimentary vector arithmetic modeling basic language constructs. The classical example of the latter is using the vector representations of the following words, results in this equality: $``king''-``man''+ ``woman'' = ``queen''$. Another simple example of a basic language construct being captured in the feature space is that the distance between the singular and plural version of words is rather consistent. The interesting part of this breakthrough, was that all of these constructs were implicitly learned. The neural network that was trained to create these embeddings, was simply fed large amounts of text and was executing prediction in 1 of 2 set-ups. These neural networks had 1 hidden layer, which is actually used as word embeddings after training.
In the first manner, termed ``CBOW'' (continuous bag of words), the model is fed a window of surrounding terms of a center term, and is tasked with predicting the center word of the window (or the target word). In the second model, termed skip-gram, the model is fed the center word, and is tasked with predicting the surrounding words. These models are essentially the reciprocal of each other, yet produce rather similar embeddings. The output of both models is a vector of probabilities for the particular prediction, that is the length of the number of unique words present in a particular training dataset.  The major pre-requisite to quality embeddings, is a large corpus to train these embeddings on. In the case of word2Vec, Mikolov et al. trained on millions of documents from Google News.  In the case of later models, they have trained on common crawl, a dataset of billions of web pages. The high level intuition with these models, is that terms used in similar contexts, likely have similar semantics. This theory is referred to as the distributional hypothesis from \citet{Harris1954}, which states that words used in similar contexts have rather similar meanings.

Since the creation of word2Vec there have been a few other notable publications on other word embedding techniques. GloVe, by \citet{Pennington}, looks to use global co-occurrence statistics combined with prediction in order to form superior word representations. Even more recently, fastText by \citet{Bojanowski2016}, which we use in our process, uses the same model as word2Vec, but instead trains at the granularity of the n-grams of words, and treats the representation of each word as a summation of the vectors of a word's n-grams.

%% file: Term_Blur_P3/term_blurring.tex

\chapter{Semantic Term ``Blurring'' }
\label{chap:term-blurring}

The common method represents a document by a row vector of term
frequencies, and represents a document corpus with a table or
matrix. The number of matrix rows is $|\mbox{D}|$, the number of the
documents in the corpus, each row corresponding to a document. The
number of matrix columns is $|\mbox{T}|$, the number of unique words or
terms used in the document corpus, each column corresponding to a
term.  The $ |\mbox{D}| \times |\mbox{T}|$ matrix is large, and
typically, it has more columns than rows, $ |\mbox{T}| > |\mbox{D}|$.
We consider in particular the case where each document is composed of
its title and abstract.  In such case, the matrix is sparse in each row.

We argue there are two issues with the document representation by term
frequencies. First, term frequencies are esoteric to an author. Certain
authors are more likely to repeat themselves to drive a point, while
others are more terse in their word usage. Second, vernacular (word
choice) is a person-specific signature as well. Each author's vernacular
is uniquely correlated to his/her upbringing, environment, training and
particular field of study. Like no two snow flakes have the same
formation pattern in detail, two authors are unlikely have the same
language “experience” and linguistic signature. Such esoteric features
are important to author identification, but pose a challenge for theme
classification of documents. 


We address the challenge with a novel approach. The basic idea is to
discount esoteric attributes among authors and explore semantic
similarity between words used by different authors. To this end, we
perform two successive operations. 
With the first operation, we convert the matrix with term counts in
integers, or frequency-based values, to a binary valued matrix of term
occurrence. Specifically, we define the occurrence matrix $\mbox{DT}$ as
follows.
\begin{equation}
\label{eq:DT-matrix} 
  \begin{array}{rl}  
  \mbox{DT}(i,j)  =  1,  &  \mbox{ if } \mbox{term}_j \in \mbox{document}_i 
  \\ 
  \mbox{DT}(i,j) =   0,  &  \mbox{ if } \mbox{term}_j \notin \mbox{document}_i 
  \end{array} 
\end{equation} 
With the second operation, we find semantic neighbors of each term and
transform the binary code for each document to a statistically
semantically weighted code. We refer to this process as semantic 
term ``blurring''. We elaborate the process in the rest of this chapter. 

To make semantic association, we perform term ``blurring''. We look to transform the feature space from
occurrences of individual terms to a weighted presence of ``semantic
elements''. Each ``semantic element'' is represented by a distribution of weights over a term neighborhood, centered at the term.

\begin{figure}[H]
    \centering
    \includegraphics[scale=0.25]{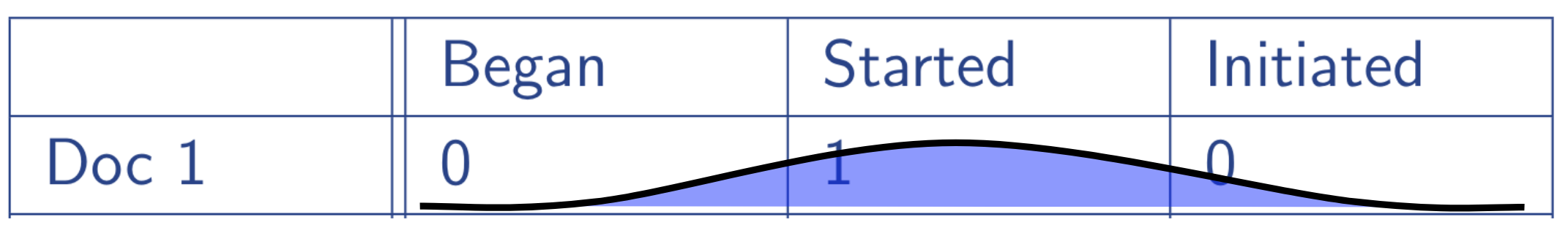} 
    \\ \includegraphics[scale=0.355]{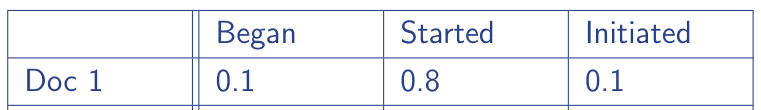}
    \caption{Illustrative Term ``Blurring'' of ``Started''}
    \label{fig:dt_transformation_gauss}
\end{figure}

\begin{figure}[H]
    \centering
    \includegraphics[scale=0.15]{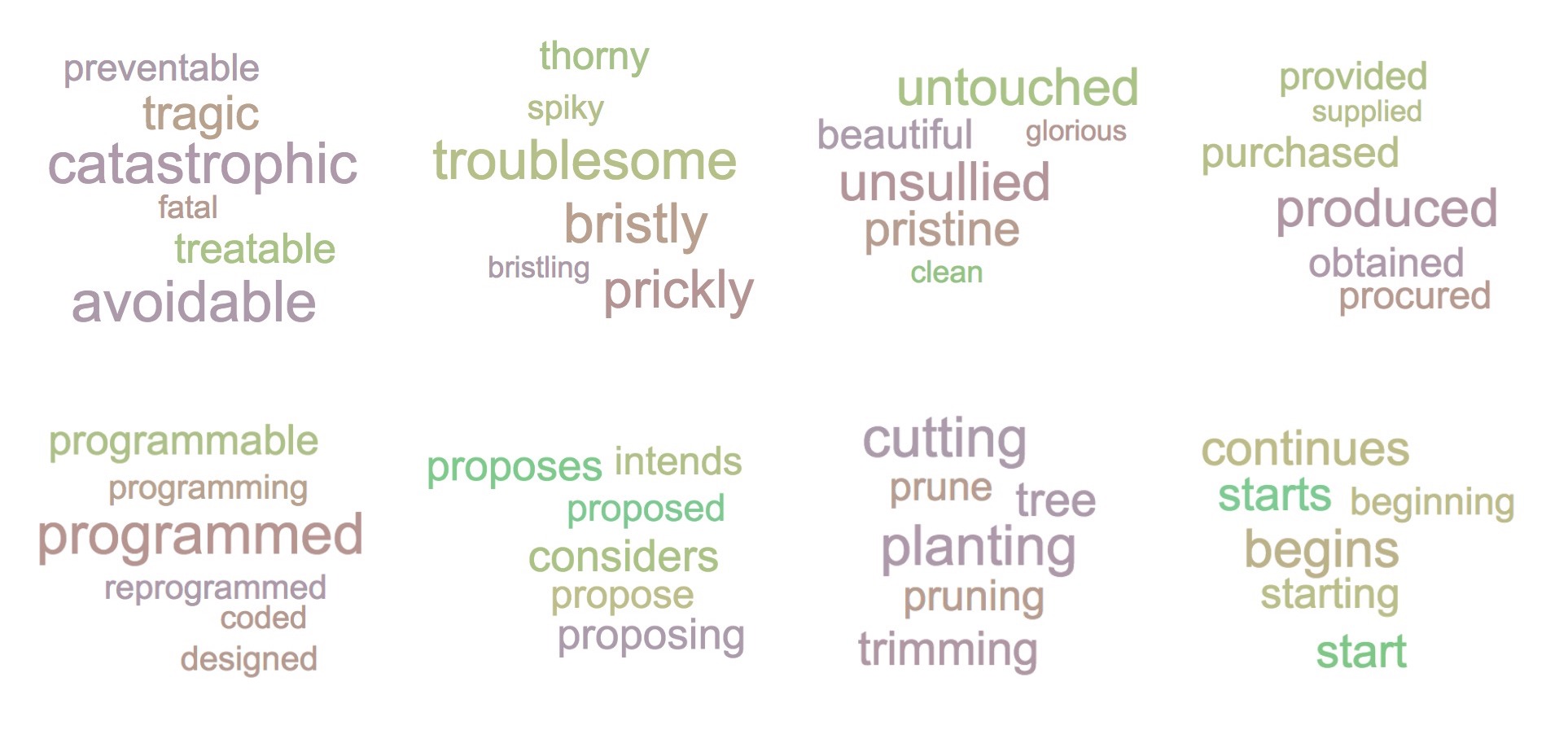} 
    \caption{Term neighbors located in the fastText \citep{Bojanowski2016} word embedding space}
    \label{fig:word_cloud_knn}
\end{figure}

The term to ``semantic element'' transformation is composed of two steps: (1) locating neighboring terms, and (2) diffusing the weights over the neighboring terms. In Figure \ref{fig:dt_transformation_gauss} the original features, representing the occurrences of ``began'', ``started'', and ``initiated'', are fully disassociated. After the transformation, the feature replacing ``started'', has weight distributed across near neighbors, but also has the potential the receive weight from neighbors as well. In other words, a ``semantic element'' is similar to a ``term neighborhood'', centered at the location of the term it replaces. ``Term neighborhoods'' are overlapping, and when this occurs weight is distributed based on distance to the center of the particular neighborhood, which is the term the ``semantic element'' is replacing. Some term neighborhoods are shown in Figure \ref{fig:word_cloud_knn}.

\section{The Need for Word Embeddings}
Our inspiration for ``term blurring'' comes from image processing, where Gaussian blurring is a common transformation, often used to increase association between different areas of an image. The challenge is that unlike images, where spatial associations and neighborhoods exist between pixels, there is not a natural equivalent for text data. Processes like Latent Semantic Analysis (LSA), have tried to capture similar terms, but struggle to do so effectively only using data from inside a particular corpus. We exploit external word embeddings to discover these near neighbor relationships between terms. Word embeddings provide us with the statistically learned spatial relationships between terms that we desire.

\section{A Bonus of Near Neighbor Term Relationships}
A classical method to attempt improving associations between terms, is called stemming. The intuition behind stemming is that different forms of a term do not warrant individual features. If we were to look at the word ``walk'', the different forms: ``walk'', ``walked'', ``walks'', and ``walking'', do not have major impact on the ``theme'' of a particular document. This is not a perfect argument, as tense may have some value, if say temporal information is useful, but for our purposes we are willing to lose that information in return for better association between documents. In the past, stemming has been done with a variety of explicit rules, such as removing a fixed list of prefixes and suffixes from terms, we look to gain the benefits of stemming from implicit near neighbor relationships.  In addition, our near neighbor relationships capture similar semantics between terms that stemming cannot (e.g. ``hi'' and ``hello'').

\section{Gaussian ``Term Blurring''}
In order to capture associations between terms with similar semantics, it requires a feature space in which near neighbor relationships imply similar semantics. This is where we utilize word embeddings. We exploit pre-trained fastText word representations in a high dimensional feature space (300-dimensions) from \citet{Bojanowski2016}, to discover term neighborhoods. More specifically, for each term in the corpus, we finds its k-nearest neighboring terms that are present in the corpus. The k-nn is calculated on the row stochastic term representations, as we are focused on correlation / similarity between terms, rather than any metric of length.  Once we find the k-nearest neighbors of a particular term, the weight is distributed across these neighbors using a Gaussian distribution. As the word embedding feature space is extremely large, we use a self-tuning $\sigma$ for our Gaussian distributions, in order to re-adjust our distributions according to the local neighborhood statistics of each ``semantic element''. The self-tuning $\sigma$ concept comes from \citet{Zelnik-Manor2005}. Unlike the cited paper, we do not use a symmetric $\sigma$, as our k-nn is not necessarily symmetric (i.e. $term_1 \in knn(term_2)$, does not imply $term_2 \in knn(term_1)$).

\begin{table}[H]
            \centering
            Before Semantic Term Blurring \\
            \begin{tabular}{ |p{2cm}||p{2cm}|p{2cm}|p{2cm}|  }
             \hline
              & Began &Started&Initiated\\
             \hline
             Doc 1   & 1    &0&   0\\
             \hline
             Doc 2&   0  & 1   &0\\
             \hline
             Doc 3 &0 & 0&  1\\
            \hline
            \end{tabular} \\
            \vspace{.5em}
            \caption{ \dtSmall Before Term ``Blurring''} 
            With the above representation: $\left \langle Doc 1,Doc 3\right \rangle = 0$ \\
\end{table}

\begin{table}[H]
            \centering
            After Semantic Term Blurring \\
            \begin{tabular}{ |p{2cm}||p{2cm}|p{2cm}|p{2cm}|  }
             \hline
              & Began &Started&Initiated\\
             \hline
             Doc 1   & 0.48    &0.34&   0.18\\
             \hline
             Doc 2&   0.33    &0.49&   0.18\\
             \hline
             Doc 3 &0.20    &0.25&   0.55\\
            \hline
            \end{tabular} \\
            \vspace{.5em}
            \caption{\dseSmall After Term ``Blurring''} 
            With the above representation: $\left \langle Doc 1,Doc 3\right \rangle \neq 0$
\end{table}

This re-weighting results in document representations that are much less sparse, increasing the associativity between features, and thus increasing the correlation between documents. In this particular trivial example, the inner product between document 1 and document 3 is initially 0. After re-weighting the inner product shows the existence of correlation. Note that this process is row sum invariant.

Formally, we again define \dtSmall to be the occurrence matrix between documents and terms, where each row represents a document and each column represents a unique term. We then define Document by Semantic Elements, \dseSmall, the result of the Gaussian ``term blurring'' below:

\begin{equation}
    \forall t_j \quad DSE(:,j) = \sum_{k: knn(t_k) \ni t_j}{{\frac{1}{s_k} DT(:,k) e^{-\frac{\left \| t_j -t_k \right \|^2}{\sigma_k^2}}}}
\end{equation}
where:
\begin{equation}
    s_i = 1+ \sum_{t_k \in knn(t_i)}e^{-\frac{\left \| t_k -t_i \right \|^2}{\sigma_i^2}}
\end{equation}
\begin{equation}
    \sigma_i = \max_{t_k \in knn(t_i)}{\left \| t_k -t_i \right \|}
\end{equation} 
Note that $knn(t_i)$ includes $t_i$. Thus $t_j \in (knn(t_k) \ni t_j).$

%% file: S_Barcoding_P4/s_barcoding.tex

\chapter{Iterative Stochastic Barcoding for
  Cluster Revision}
\label{chap:s-barcoding}

The second contribution we made is an iterative cluster revision method
using what we deem as ``stochastic barcoding''.  More often than not,
clustering is carried out in a dimension-reduced space, largely due to
the fact that the data points (documents in our case) are extremely
sparse in the high dimensional feature space. In a low dimensional feature
space, certain properties are preserved, while other properties are lost
or distorted. For example, in an embedding space obtained by t-SNE, only
stochastic neighborhoods are to be preserved.  We obtain a cluster
configuration in a low-dimensional embedding space, and we revise the cluster configuration in the high-dimensional space, i.e., revising cluster memberships as well
as the total number of clusters. We also exploit both geometric and
stochastic relationships within and between clusters.

A closely related previous work is by Dhillon et al in
2002\cite{Dhillon2002}~\footnote{The reference was found and pointed
  out to us by Alexandros Illiopoulos upon his learning of our
  ``S-Barcoding'' method.}.  The authors attempt to locate k-clusters in the
high dimensional space with the spherical k-means algorithm, and when the iterative
k-means search algorithm gets stuck at a local maximum, they deploy a
refining strategy, named first variation, to pull the iteration process
out of the local maximum. The clusters are located and refined in the same
feature space, by the same metric, the same objective function, and by
the same number of clusters. Computationally, the algorithm extremely
sequential, letting one document migrate from one cluster to another at
a time.

\section{Our Process}

\begin{figure}[H]
    \centering
    \includegraphics[scale=0.15]{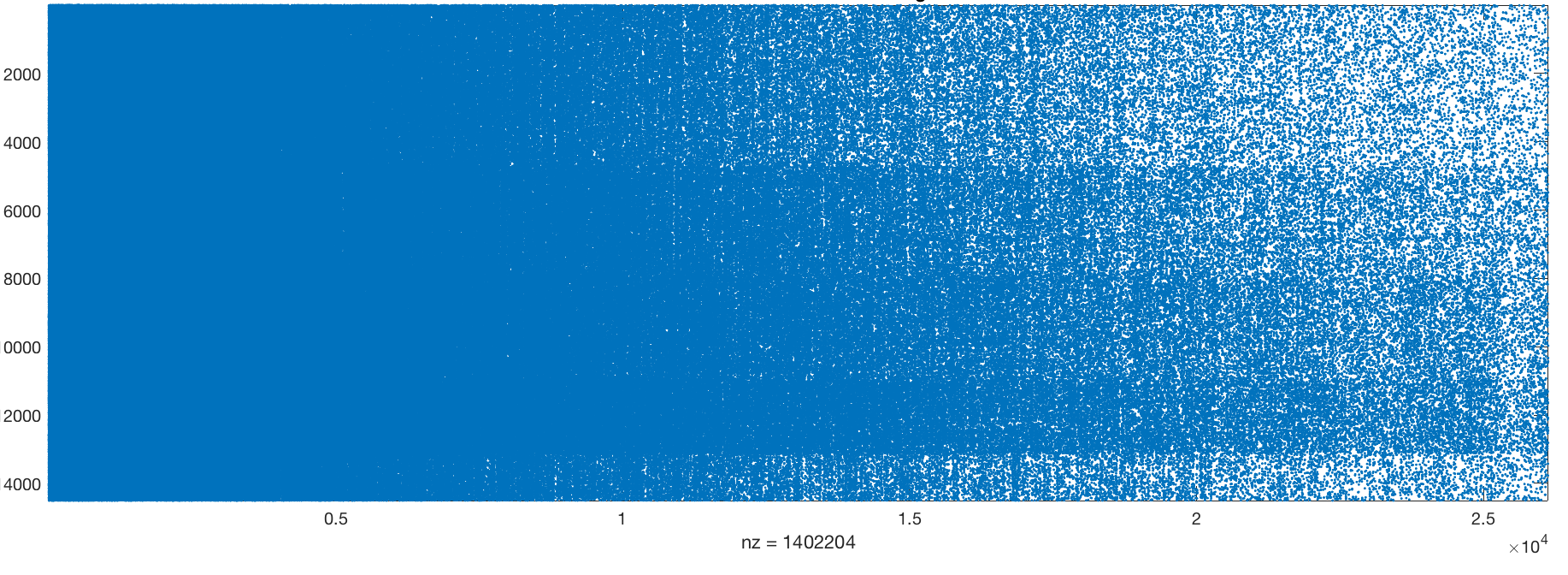} 
    \caption{Ground Truth \dseSmall, documents of the same theme 
    are permuted into the same band of rows, five bands correspond 
    to five themes, each band has a unit stochastic SE pattern. 
    } 
    \label{fig:barcodeGT}
\end{figure}

We integrate structures and information from both low and high
dimensional spaces. We obtain first a cluster configuration in a low
dimensional embedding space. We revise the cluster configuration in the
high-dimensional feature space. Our initial intuition came from visual
observation of the statistical patterns of semantic elements in each
cluster and between clusters. In Figure \ref{fig:barcodeGT}, we show the \dseSmall
matrix, with documents in the same cluster put in the same band of rows.

In this ordering, 5 ``bands'' of rows are clearly distinguishable from each
other. These bands correspond to the 5 true labels of the particular
corpus. The vertical bars of each band manifest the stochastic patterns of
the semantic elements unique to the cluster. We therefore refer to the
the patterns and their numerical values as stochastic barcodes
(``S-Barcodes").

By cluster revision, we look to uncover or get closer to the stochastic
barcodes for the ground truth clusters. We start with the barcodes for the initial
clusters rendered by a a clustering algorithm from a low-dimensional
representation.
For each cluster, we give its ``S-Barcode'' a vector of numerical values 
as follows. 
\begin{equation} 
\label{eq:S-barcode} 
\text{\barcode}(i,:) = \frac{\underset{d_k \in D_i}{\sum}{\text{\dseSmall}(k,:)}}{\left \| \underset{d_k \in D_i}{\sum}{\text{\dseSmall}(k,:)} \right \|_1}
\end{equation} 
We then re-examine and revise cluster membership, inter-cluster relationships, intra-cluster relationships, and the number of clusters.

\section{Cluster Revision}

First, perform a dimension reduction on the text data, then cluster in the lower dimensional space. To revise, for each document cluster from the low dimensional clustering, form an “S-Barcode” for the cluster, which is defined in \ref{eq:S-barcode} Then iterate through each document, and re-assign the document to the cluster the ``S-Barcode'' of which  it is most correlated, or in other words a document is linked to the ``S-Barcode'' for which is has the maximal dot product. This process is repeated until convergence.\\
\textbf{The S-Barcoding Algorithm}\\
Initialize with clusters from low-dimensional space, repeat until
convergence:
\begin{itemize}
     \item For each synthetic document class, $D_i$, calculate its ``S-Barcode'' by \ref{eq:S-barcode}
     \item For each document $d_i$ find the barcode with the maximum correlation
     \begin{itemize}
         \item $d_i$ is assigned to $\displaystyle  \underset{j}{\arg\max}{\frac{d_i^T \text{\barcode}(j)}{\left \| d_i \right \|_2 \left \| \text{\barcode}(j) \right \|_2}}$
     \end{itemize}
 \end{itemize}
In our experiments, the iteration reaches stationary configuration in a few steps.

For implementation of this method, simply multiply \dseSmall (or \dtSmall), by the transpose of the barcodes and then find the maximum value in each row. In \matlab can be expressed as: \mbox{max($\text{DSE} \cdot \text{\barcode}^T$,[],2)}.


%% file: Experimental_results_P5/experimental_results.tex
\chapter{Experimental Results}
To experimentally evaluate our new ``blurred'' document representations with semantic term blurring and cluster revision with stochastic barcoding, we developed an architecture for the aggregation and clustering of academic documents, as elaborated on by \citet{david2018}. We have the option for forming document representations by term frequencies or semantic element representations with semantic ``term blurring''. We then perform a dimension reduction on the documents, using either a geometric dimension reduction method (SVD) or a statistical method (t-SNE). We cluster the documents in the reduced dimension space using mean shift or K-means. We perform cluster revisions in the original high-dimensional space utilizing ``stochastic barcoding''.

For our experiments we compare performing the unsupervised classification process with: the original term frequency representations, \dtSmallNoS, and then with our re-weighted semantic element representations, \dseSmallNoS. We do so on two separate corpora. In our results we are able to obtain improved cluster purity and overall clustering quality when utilizing our new document representations after ``term blurring". Futhermore, we see a significant improvement in clustering results after using our iterative ``S-Barcoding'' method. 

For both \dtSmall and \dseSmall matrices, we use a high and low pass filter. In the case of \dtSmallNoS, we remove terms have been used in one or no documents throughout the corpus. In the case of \dseSmallNoS, we remove features where the column sum is 1 or less, but with \dseSmallNoS, these feature values weighted, so it does not necessarily correspond to a term being used once. After this high pass filter, there are about an equivalent number of features remaining for both \dtSmall and \dseSmallNoS, despite different weights and interpretations. For the low pass filter, for both matrices, we remove all columns where the column sum is in the top $95\%$ of all column sums.

For \dseSmall, $k = 4$ for all experiments, meaning for each term, weight is distributed across its 4 nearest neighbors using the weighting scheme described earlier.

\section{Corpus 1}

\subsection{Composition}
The first corpus consists of 14,502 documents spread out across five particular topics.
The breakdown of the documents is as follows:
\begin{itemize}
    \item 3219 Brain Cancer
    \item 4707 Computer Vision
    \item 2154 Ecology
    \item 1376 Music
    \item 3046 Physical Activity
\end{itemize}
This corpus results in a $DT$ matrix of size $14502 \times 26,990$ after filtering.

\subsection{Low Dimensional Embedding}

\begin{figure}[H]
\begin{minipage}{.5\textwidth}
  \centering
 DT Affinity
  \includegraphics[width=1\linewidth]{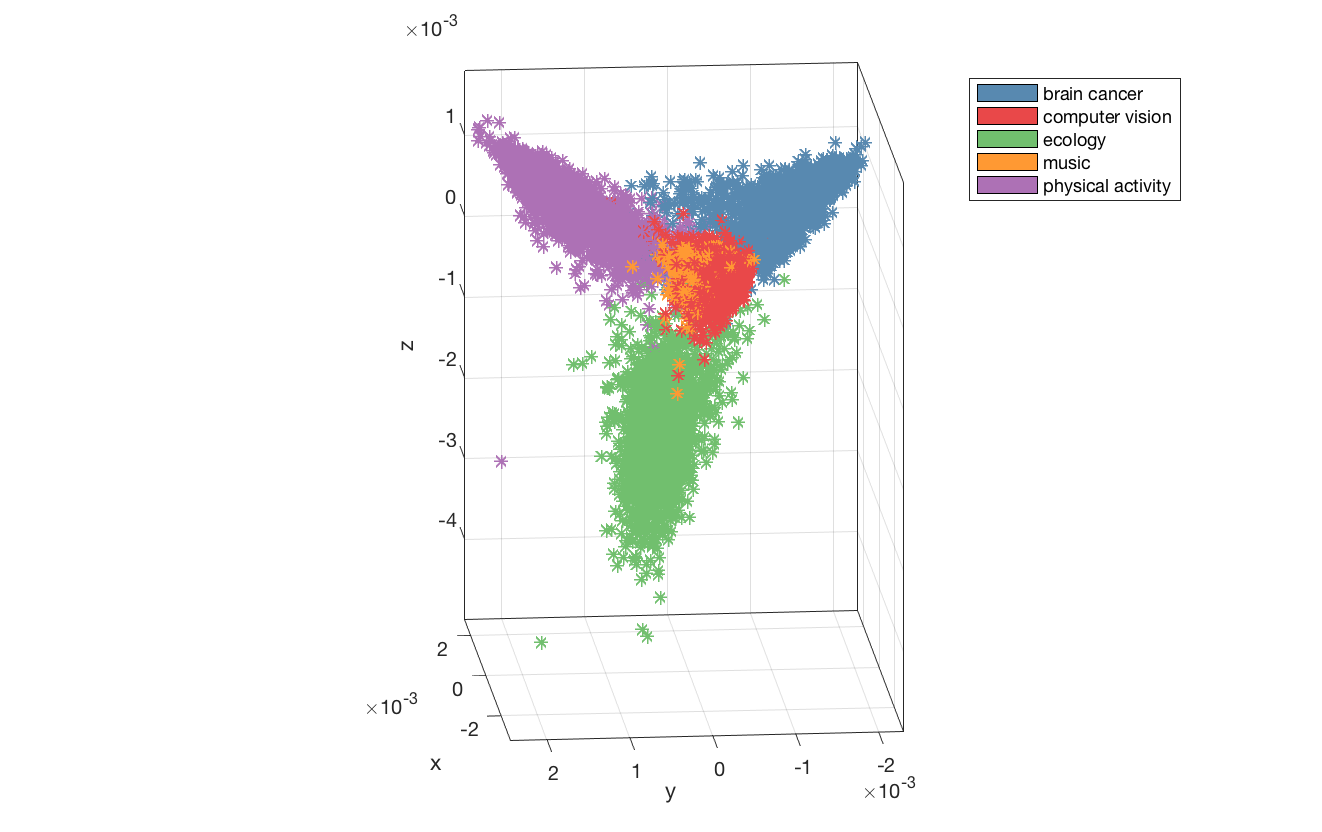}
\end{minipage}%
\begin{minipage}{.5\textwidth}
  \centering
  DSE Affinity
  \includegraphics[width=1\linewidth]{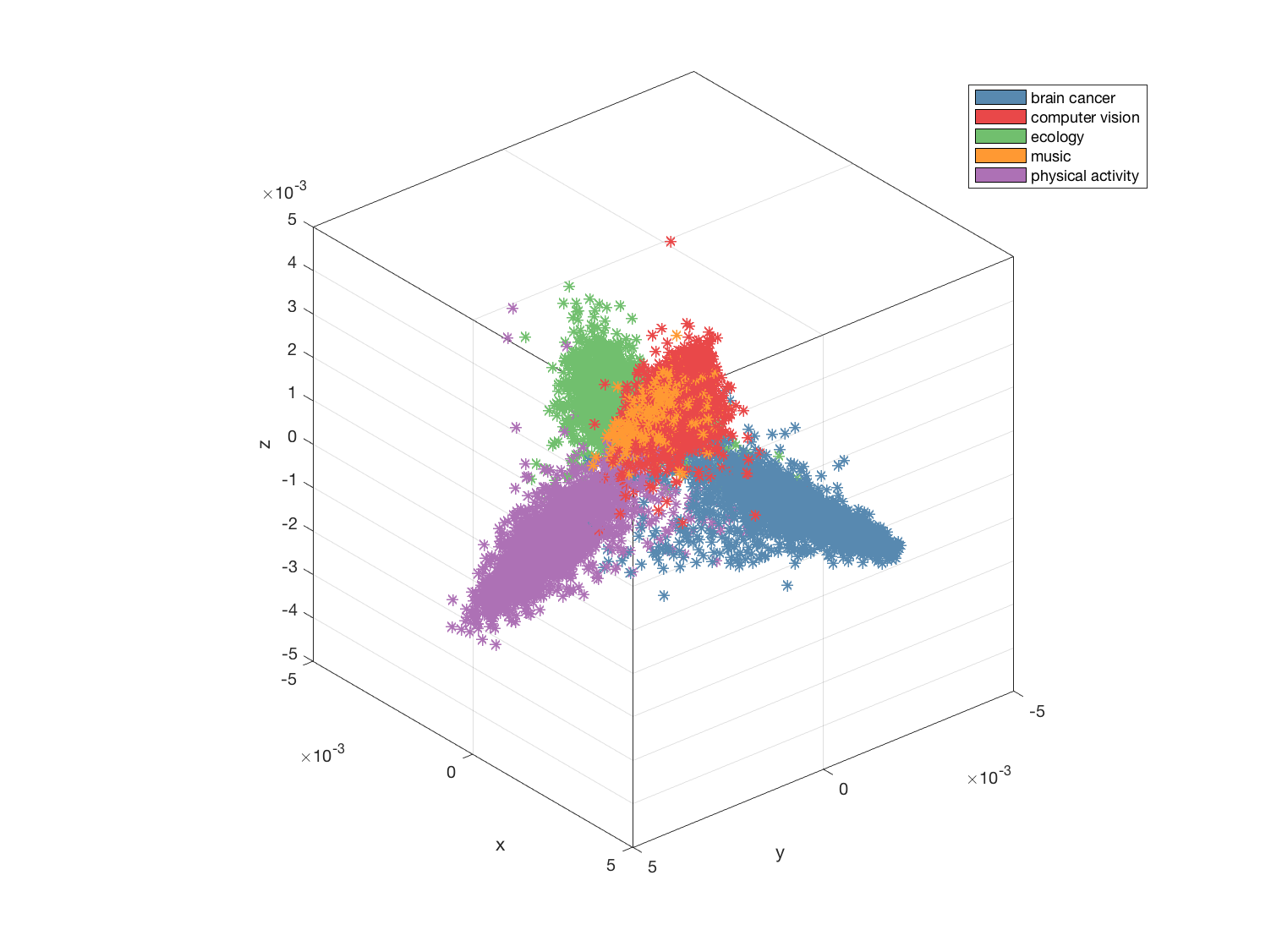}
\end{minipage}
\caption{Document points (corpus 1) in 3-dimensional space via bipartie spectral embedding. (L) from DT affinity matrix, (R) from DSE affinity matrix.}
\end{figure}
This corpus, for the most part, is about 5 pretty disparate topics, which can be perceived in the 3-dimensional spectral embedding and encoding. However, while slightly difficult to see in the 3-dimensional embedding, the DSE affinity embedding has superior separation for the music cluster from the rest, in comparison to DT, where music is dispersed throughout the computer vision papers. This is very clear in the clustering results.

\subsection{Clustering Analysis}
\begin{figure}[H]
\begin{minipage}{.5\textwidth}
  \centering
  Confusion Matrix (DT Affinity)
  \includegraphics[width=1\linewidth]{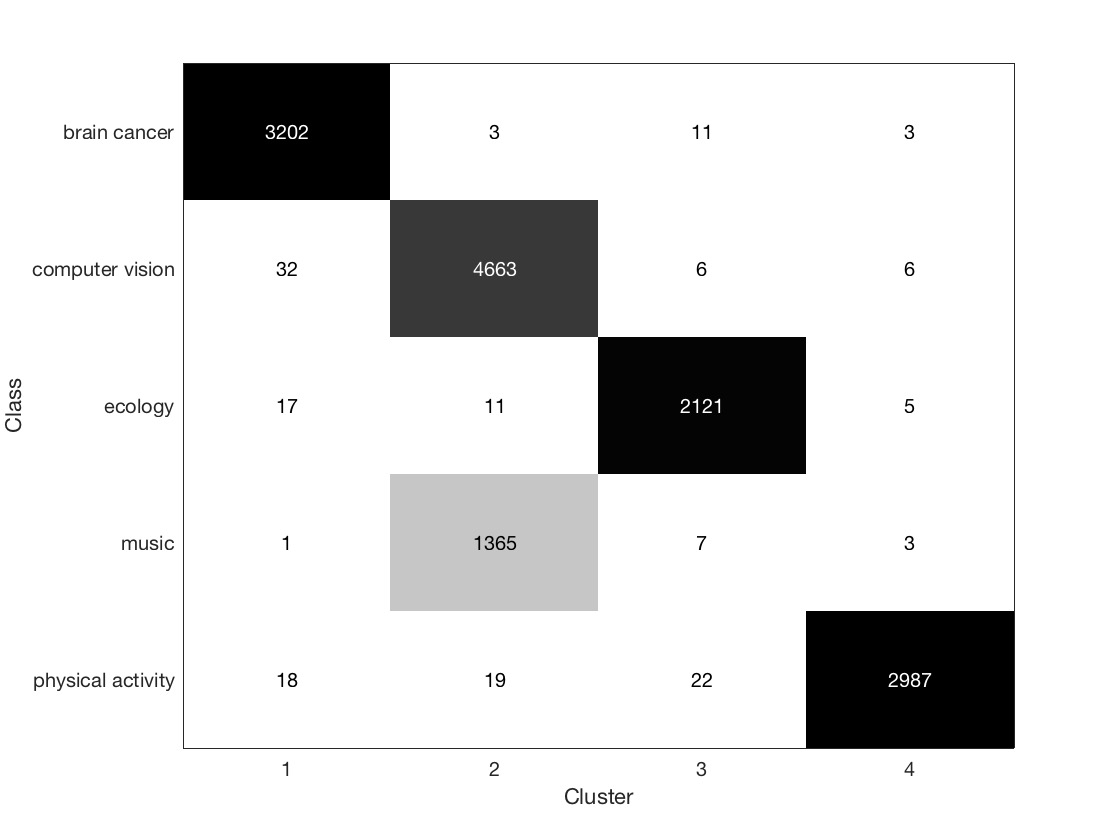}
\end{minipage}%
\begin{minipage}{.5\textwidth}
  \centering
  Confusion Matrix (DSE Affinity)
  \includegraphics[width=1\linewidth]{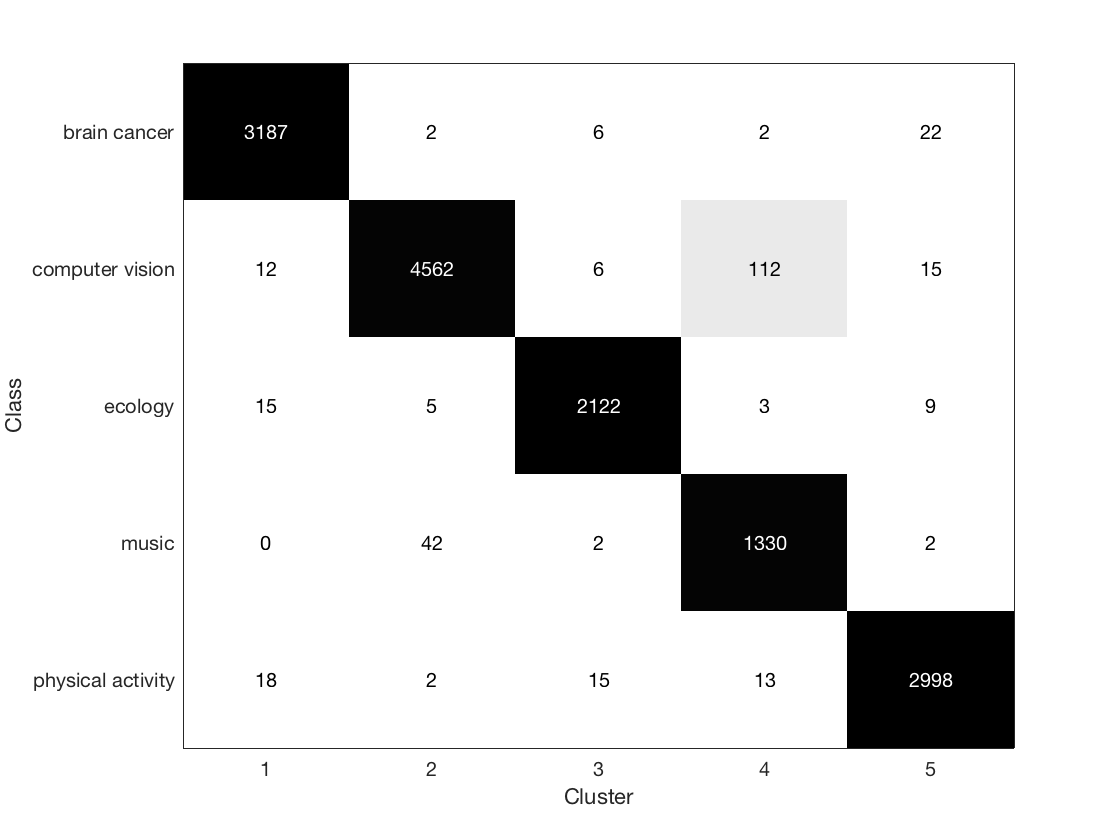}
\end{minipage}
\caption{Document clustering results (corpus 1), using the (L) DT affinity  and (R) DSE affinity, after ``S-Barcoding''. Both were clustered using mean shift with $\sigma = .001$ in a 5-dimensional embedding space, via spectral bipartite embedding. This is then followed by ``S-Barcoding'', with a minimum cluster size of 5. Prior to ``S-Barcoding'', mean shift results in 16 initial clusters for the DT Affinity spectral embedding and finishes with 4 clusters after ``S-Barcoding'' (L). For the DSE Affinity (R), mean shift results in 15 clusters for the DSE Affinity spectral embedding and finishes with 5 clusters after ``S-Barcoding'', which directly matches the number of true labels in the corpus. Note that the music class is missed entirely on the left (DT Affinity), but is captured on the right (DSE Affinity). The ``purity'' of the clusters before and after revision is comperable for both the DT and DSE Affinity clustering results.}
\end{figure}

\subsection{``S-Barcode" Revision Results}
\begin{figure}[H]
\begin{center}
    ``S-Barcodes" For 15 Clusters Before Revision
    \includegraphics[scale=.15]{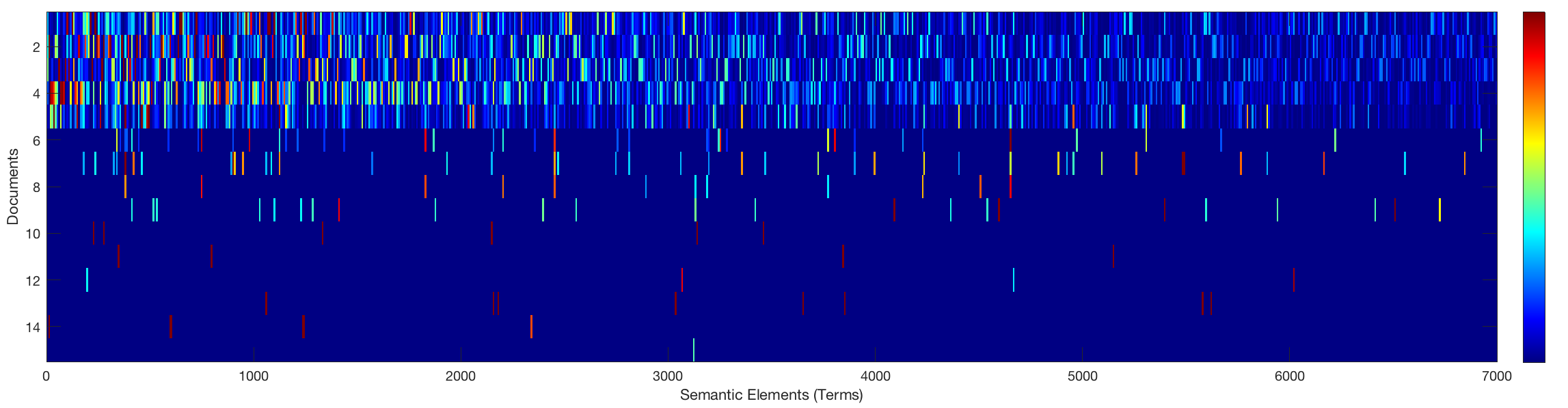}
\end{center}

\begin{center}
    ``S-Barcodes" For 5 Clusters After Revision

    \includegraphics[scale=.15]{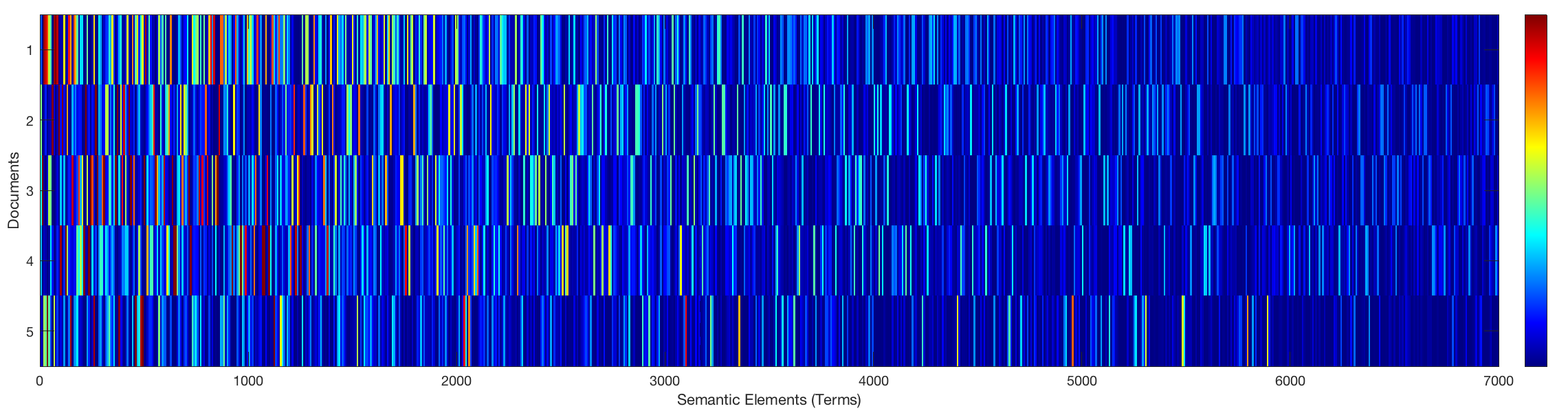}
\end{center}
\caption{The ``S-Barcodes'' before and after cluster revision (corpus 1). What is shown is the leading and dominant portion of the barcode bands (showing the most significant 7000 features). Both the lower and higher dimensional feature spaces using the \dseSmall affinity; lower dimensional embedding via spectral bipartite SVD embedding. Clustering in the lower dimensional space (5D), completed using mean shift with $\sigma = .001$. ``S-Barcoding'' executed with minimum cluster size of 5.}
\end{figure}

\begin{figure}[H]
\begin{minipage}{.5\textwidth}
  \centering
  Confusion Matrix Before ``S-Barcode" Revision \\
  \vspace{1em}
  \includegraphics[scale=.1]{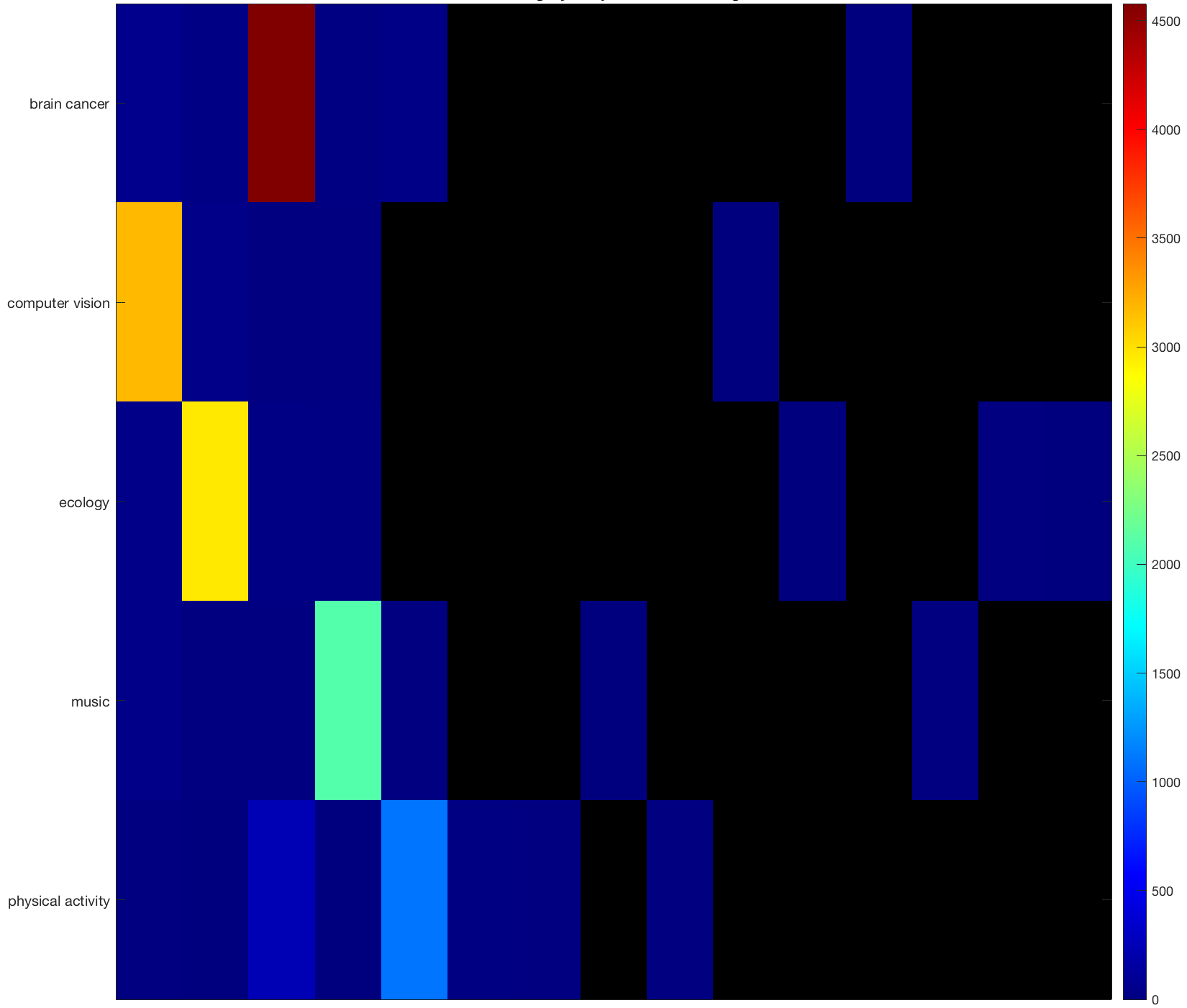} \\
\end{minipage}%
\begin{minipage}{.5\textwidth}
  \centering
  Confusion Matrix After ``S-Barcode" Revision \\
  \includegraphics[width=1\linewidth]{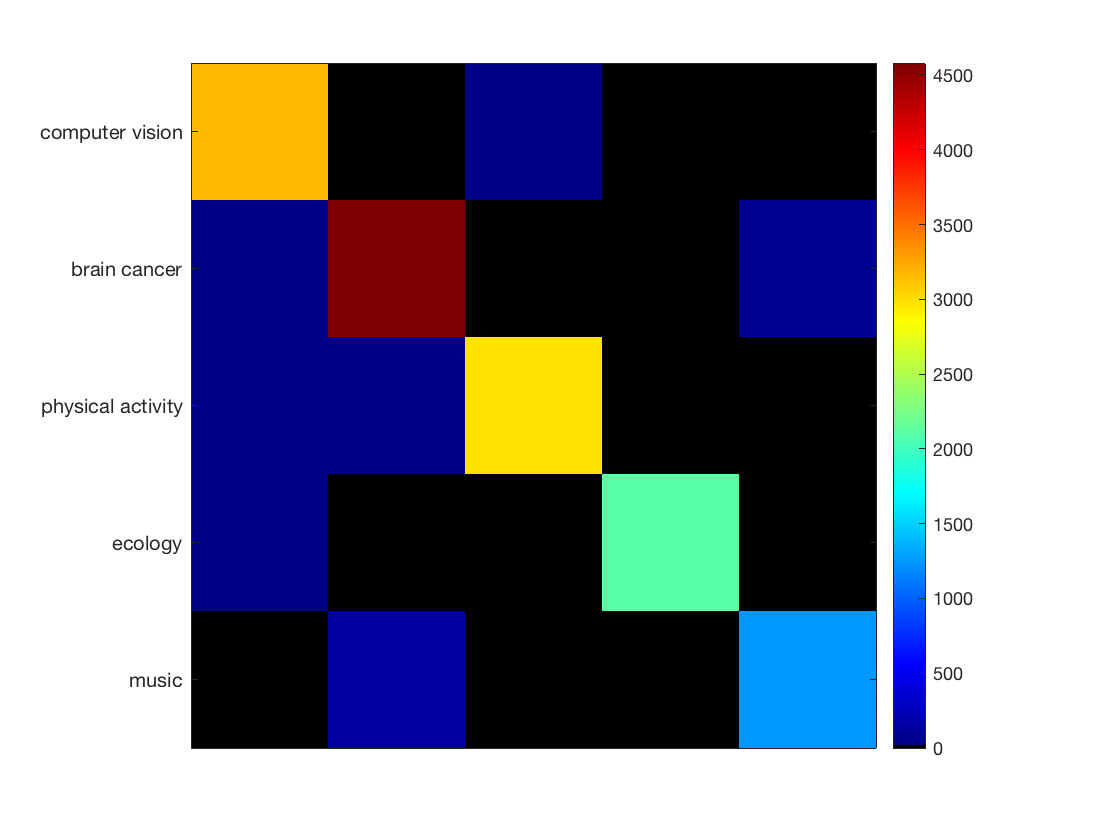}
\end{minipage}
\caption{Document clustering results (corpus 1), before ``S-Barcoding'' (L) and after ``S-Barcoding'' (R). These results are both on the \dseSmall Affinity. In these plots, color represents number of documents in a cluster.}
\end{figure}

\section{Corpus 2}
\subsection{Composition}
The second corpus is the six largest categories from the popular Reuters News dataset \citep{ReutersData}, limited only to documents containing at least 50 words. This thresholding on minimal documented length narrows the dataset to 6818 documents. 
The breakdown of the documents is as follows:
\begin{itemize} 
  \item 2098 Mergers/Acquisitions, or `acq'
\item 335 Crude oil, or `crude'
\item 3670 Earnings/Earnings forecasts, or `earn'
\item 182 Interest rates, or `interest'
\item 231 Money/Foreign Exchange, or `money-fx'
\item 302 Trade news, or `trade' 
\end{itemize}
This corpus results in a $DT$ matrix of size $6818 \times 17,546$ after filtering.
\subsection{Clustering Analysis}
\begin{figure}[H]
  \begin{minipage}{.5\textwidth}
    \centering
    Confusion Matrix (DT Affinity)
    \includegraphics[width=1\linewidth]{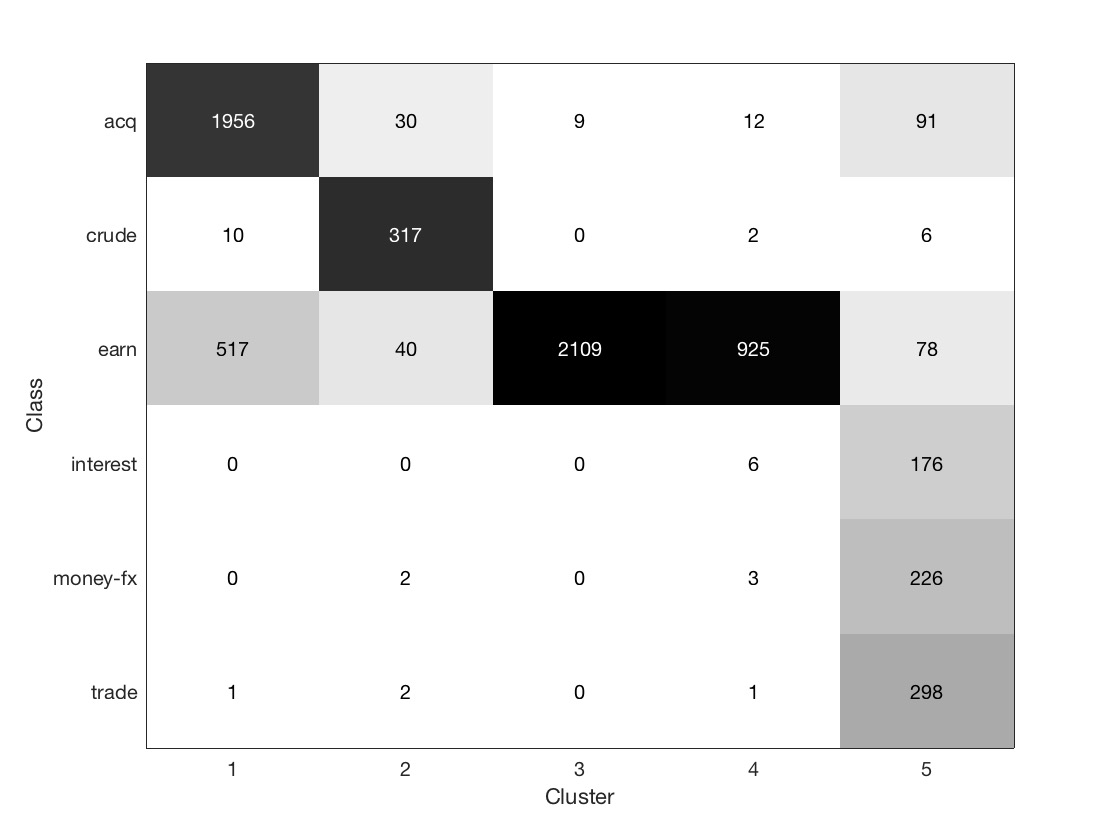}
  \end{minipage}%
  \begin{minipage}{.5\textwidth}
    \centering
    Confusion Matrix (DSE Affinity)
    \includegraphics[width=1\linewidth]{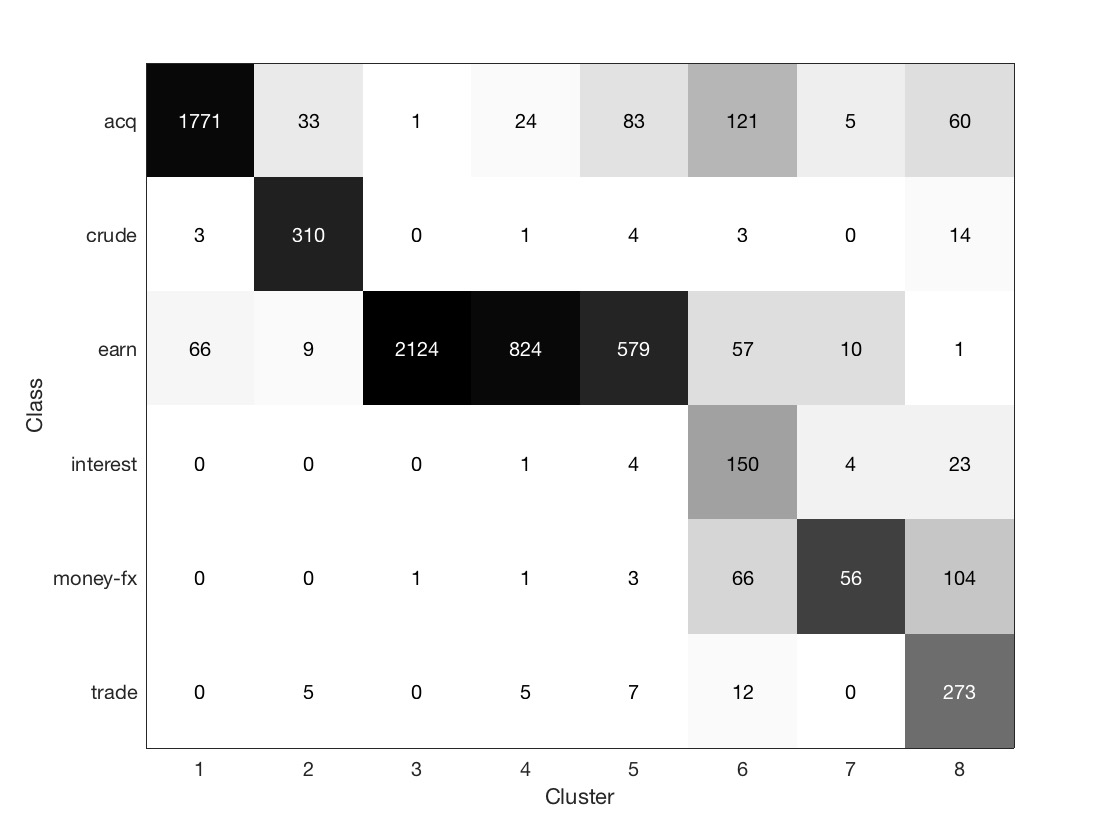}
  \end{minipage}
  \caption{Document clustering results (corpus 2), using the (L) \dtSmall affinity  and (R) \dseSmall affinity. Both were clustered using mean shift in a 10-dimensional embedding space, with: $\sigma = .003$ for \dtSmall Affinity and $\sigma = .00425$ for \dseSmall Affinity. The clustering results presented above are after ``S-Barcoding''revision, with a minimum cluster size of 10. Prior to ``S-Barcoding'', mean shift results in 55 initial clusters for the \dtSmall Affinity spectral embedding and finishes with 5 clusters after ``S-Barcoding'' (L). The cluster purity before barcoding was $69\%$, and after barcoding is $82\%$.  For the \dseSmall Affinity (R), mean shift results in 54 clusters for the \dseSmall Affinity spectral embedding and finishes with 8 clusters after ``S-Barcoding''. The cluster purity before ``S-Barcoding'' is $72\%$, and is $89\%$ after ``S-Barcoding''.}
  \end{figure}


\subsection{``S-Barcode'' Revision Results}
\begin{figure}[H]
  \begin{minipage}{.5\textwidth}
    \centering
    Confusion Matrix (DSE Affinity): K-means

    \includegraphics[width=1\linewidth]{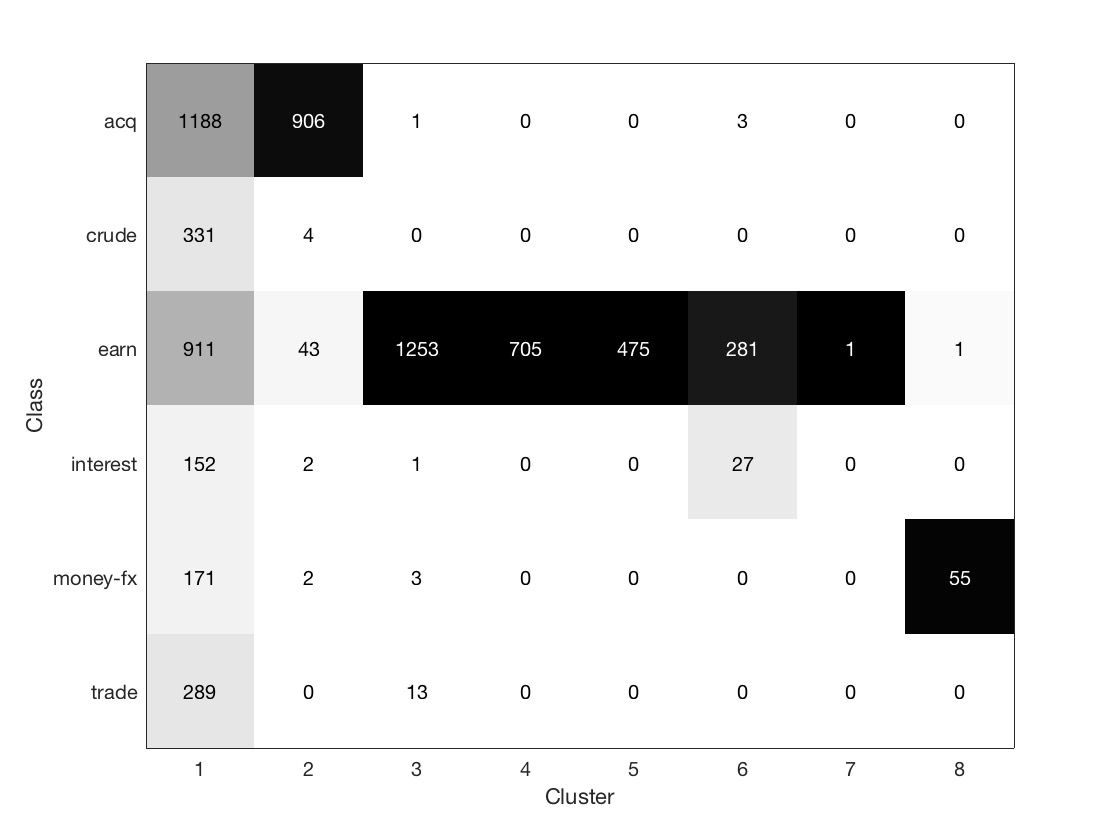}
  \end{minipage}%
  \begin{minipage}{.5\textwidth}
    \centering
    Confusion Matrix (DSE Affinity): Mean shift and ``S-Barcoding''
    \includegraphics[width=1\linewidth]{Pictures/reuters_dse_confmat22.jpeg}
  \end{minipage}
  \caption{Comparison of document clustering results (corpus 2), using the DSE Affinity, clustering with (L) K-means  and (R) mean shift followed by ``S-Barcoding''. K-means is executed with $K=8$, to compare with ``S-Barcoding'', which resulted in 8 clusters. For (R) mean shift was run with $\sigma = .00425$ on 10-dimensional spectral bipartite embedding. Following this,  both were in a 10-dimensional embedding space, followed by ``S-Barcoding'', with a minimum cluster size of 10. Prior to ``S-Barcoding'', mean shift results in 54 initial clusters in the low dimensional space and results in 5 clusters after ``S-Barcoding''. The cluster purity before ``S-Barcoding'' is $72\%$, and is $89\%$ after ``S-Barcoding'', for the initial mean shift results (R). The K-means purity (L) is $69\%$.}
  \end{figure}

In these results, we see superior clustering when utilizing mean shift and then using ``S-Barcoding'' revision. The mean shift version used is not optimal, as it assumes a single Gaussian mode. Thus, these results are a proof of validity, but likely could be improved significantly, by parameter tuning or a better clustering algorithm.


%% file: Discussion/discussion.tex
\chapter{Discussion}
One area for potential improvement involves the dividing and  merging of clusters during ``S-Barcoding''. Currently, documents are assigned based on correlation to their most correlated cluster. This does not take into account the ``shape'' of the cluster, or if clusters begin to overlap. We are exploring methods for cluster merging in the high dimensional space.

The next area for improvement, also related to ``S-Barcoding'', involves identifying the features that either discern between clusters or associate two clusters. Assume we have two clusters about different themes, $C_1$ and $C_2$. We aim to find the features (either ``semantic elements'' or ``terms''), that discern $C_1$ from $C_2$. Now imagine we have a third cluster, $C_3$, where $C_3$ is about the same theme high level theme as $C_1$, but relates to a different subset of that theme. For $C_1$ and $C_3$ we would like to find the features that 1) indicate the clusters are about the same general theme, but also the features that 2) indicate the different subset of the theme $C_1$ and $C_3$ describe.

A third note is that in the future we would like to make more quantitative comparisons between word embedding models. No strong evaluation metrics exist for word embedding models, thus often downstream results are used to evaluate the quality of word embeddings. Throughout the project we used word2Vec \citep{Mikolov2013}, GloVe \citep{Pennington}, and fastText \citep{Bojanowski2016} models and the different models did have significant impact on our downstream clustering results. We would like to further analyze and quantify these differences.

%% file: Biography/biography.tex
\biography
\begin{itemize}
\normalbaselines
\item Robert Frank Martorano III
\item Birthday: January 26, 1996 (Hackensack, NJ)
\item B.S. Computer Science from Duke University
\item After graduation working at Verily Life Sciences (formally Google[x]) as a Software Engineer on their clinical studies platform.
\end{itemize}
